\documentclass[10pt,journal,twocolumn,oneside]{IEEEtran}

\IEEEoverridecommandlockouts

\usepackage[utf8]{inputenc}          

\usepackage{amsmath,amsfonts,amssymb,amsthm,bm}

\usepackage{graphicx}                
\usepackage[dvipsnames]{xcolor}      
\DeclareGraphicsExtensions{.pdf,.png,.jpg,.eps}

\usepackage{caption}
\usepackage{subcaption}
\usepackage{float}
\usepackage{algorithm}
\usepackage{algpseudocode}

\usepackage{csvsimple}
\usepackage{enumerate}
\usepackage{enumitem}
\setlist{nolistsep}

\usepackage{setspace}
\usepackage{comment}
\usepackage{soul}
\usepackage{xspace}

\usepackage{listings}
\usepackage[export]{adjustbox}

\usepackage{optidef}

\usepackage[numbers,sort&compress]{natbib}
\usepackage{hyperref}
\hypersetup{
    colorlinks=true,
    linkcolor=MidnightBlue,
    citecolor=MidnightBlue,
    urlcolor=MidnightBlue
}

\begin{document}

\title{\LARGE \bf 
{Real-Time Performance Analysis of Multi-Fidelity Residual Physics-Informed Neural Process-Based State Estimation \\ for Robotic Systems\\
}}
\author{Devin Hunter and Chinwendu Enyioha
\thanks{The authors are with the Electrical \& Computer Engineering Department at the University of Central Florida, Orlando FL 32816, USA. Emails: \texttt{de700090@ucf.edu, cenyioha@ucf.edu}. The research was funded, in part, by the Florida Educational Fund and McKnight Foundation.}
}

\maketitle

\begin{abstract}

Various neural network architectures are used in many of the state-of-the-art approaches for real-time nonlinear state estimation. With the ever-increasing incorporation of these data-driven models into the estimation domain, model predictions with reliable margins of error are a requirement -- especially for safety-critical applications. This paper discusses the application of a novel real-time, data-driven estimation approach based on the multi-fidelity residual physics-informed neural process (MFR-PINP) toward the real-time state estimation of a robotic system. 
Specifically, we address the model-mismatch issue of selecting an accurate kinematic model by tasking the MFR-PINP to also learn the residuals between simple, low-fidelity predictions and complex, high-fidelity ground-truth dynamics. To account for model uncertainty present in a physical implementation, robust uncertainty guarantees from the split conformal (SC) prediction framework are modeled in the training and inference paradigms. We provide implementation details of our MFR-PINP-based estimator for a hybrid online learning setting to validate our model's usage in real-time applications. Experimental results of our approach's performance in comparison to the state-of-the-art variants of the Kalman filter (i.e. unscented Kalman filter and deep Kalman filter) in estimation scenarios showed promising results for the MFR-PINP model as a viable option in real-time estimation tasks.  
\end{abstract}

\section{Introduction and Background}

Accurate and reliable state estimation remains a cornerstone of autonomous robotic systems, directly impacting their ability to localize, navigate, and interact with complex environments. In recent years, a growing body of research has explored both model-based and data-driven paradigms to achieve robust estimation under challenging sensing conditions \cite{feng2023hybridstateest,Housein2022EKF,jin2021new}. At one end of the spectrum, dead reckoning approaches relying on motion-based sensors such as inertial measurement units (IMUs) and wheel encoders provide a lightweight and fully onboard solution for estimating a ground robot's odometry. Such methods are appealing due to their low computational and sensor requirements, making them attractive in cost- and weight-constrained scenarios. However, it is well established that dead reckoning accumulates drift over time, with errors in wheel encoder calibration, slippage, or IMU bias. These errors result in large-scale localization inaccuracies when external correction signals such as GPS, LiDAR, or vision are unavailable \cite{niu2021wheel,Borenstein1995UMBmark}.

To overcome these limitations, model-based sensor fusion techniques such as the Extended Kalman Filter (EKF), Unscented Kalman Filter (UKF), and factor graph optimization methods have been widely deployed \cite{moore2015generalized,feng2023review}. These approaches combine data from multiple sensors including LiDAR, cameras, IMUs, and encoders, while explicitly encoding kinematic and dynamic models of the platform. Although successful in practice, they require accurate motion models and well-calibrated sensors. In addition, performance can degrade significantly under sensor failures or mismatches between assumed and true system dynamics.  

In parallel, the growth of machine learning has motivated data-driven state estimation methods, such as deep Kalman filters, recurrent neural networks, and more recently, attention-based sequence models. These methods bypass explicit modeling by learning latent state representations directly from sequential data. While capable of capturing complex nonlinearities, they often require large volumes of labeled training data, utilized models can output physically \emph{infeasible} predictions, and they typically lack principled mechanisms for quantifying predictive uncertainty. This makes them difficult to deploy in safety-critical robotic applications, where predictions need to be physically consistent with the system being modeled, and confidence bounds on estimates are as important as the estimates themselves.  

Data-driven estimators such as deep Kalman filters\cite{krishnan2015deep,revach2022kalmannet}, recurrent networks \cite{doerr2018probabilistic}, and sequence models learn latent dynamics directly from data and can capture complex nonlinearities that elude hand-tuned filters \cite{goel2024can}. However, many such approaches require large labeled datasets and often provide limited, poorly calibrated uncertainty estimates, reducing their appeal for safety-critical robotic systems. Neural Processes (NPs) \cite{dubois2020npf,garnelo2018neuralprocesses} were introduced as a family of models that combine the flexibility of neural nets with the fast adaptation and uncertainty quantification of stochastic processes; NPs condition on context sets and output predictive distributions, making them attractive for few-shot and online adaptation tasks. Attentive Neural Processes (ANPs) extended NPs by adding attention mechanisms to overcome underfitting and to produce query-specific context representations, substantially improving predictive fidelity \cite{kim2019attentive}.

Recently, a promising line of work has focused on multi-fidelity state estimation methods. Multi-fidelity approaches leverage complementary information from both low-fidelity models, such as simplified kinematics, inexpensive simulators, or approximate analytical models, and high-fidelity data sources, such as real sensor logs, physics-based simulators, or ground truth from motion capture \cite{pawar2022pgmlmultifidelity,Giselle_Fern_ndez_Godino_2023,Wu2022MFHNP}. By exploiting multiple fidelity levels, such approaches can achieve both data efficiency and generalization, using the lower-fidelity models to guide learning of highest fidelity high-fidelity corrections to ensure accuracy. Within this space, Gaussian processes, Bayesian hierarchical models, and neural operators have all been investigated as multi-fidelity surrogates \cite{ravi2024multi,lu2022multifidelity,do2023multi}. Yet, most existing methods remain constrained by scalability challenges, limited real-time applicability, and weak handling of structured uncertainty \cite{fernandez_godino2023review,Wu2022MFHNP,cutajar2019deep}.  

These challenges motivate the need for a state estimator that operates effectively in dead-reckoning settings with minimal sensor suites, leverages multi-fidelity modeling to compensate for imperfect kinematic priors, and provides robust and adaptive uncertainty quantification suitable for real-time robotic deployment.  

In this work, we propose the Multi-Fidelity Residual Physics-Informed Neural Process (MFR-PINP), a neural process-based state estimation framework designed to address these challenges. The MFR-PINP extends the work in \cite{hunter2025hybrid} by introducing a residual learning mechanism that explicitly models the discrepancy between simple, low-fidelity predictions and complex, high-fidelity ground-truth dynamics. This residual formulation enables the estimator to correct systematic biases introduced by approximate kinematic models while still benefiting from the inductive structure they provide. Furthermore, by embedding conformal prediction into the training and inference loop, the MFR-PINP delivers calibrated uncertainty sets that adapt to changes in real-world sensor behavior. 

The contributions of this work are twofold. 
\begin{itemize}
    \item First, we introduce the MFR-PINP, a real-time, multi-fidelity residual learning framework that fuses physics-based low-fidelity priors with high-fidelity observational data for state estimation.
    \item Second, we validate the proposed approach in a hybrid online learning setting, demonstrating its effectiveness in GPS-denied environments using only low-fidelity motion data at inference, while benchmarking against state-of-the-art baselines.
\end{itemize}

By combining multi-fidelity modeling, physics-informed residual learning, and principled uncertainty calibration, the MFR-PINP establishes a novel approach for deploying real-time state estimation solutions onto autonomous robotic systems operating in challenging environments.  

The remainder of this paper is structured as follows. In Section \ref{sec:problem-statement}, we formally present the problem. In sections \ref{sec:method} and \ref{sec:results}, we present the approach and outcomes, respectively. We make concluding remarks in Section \ref{sec:conclude+future}.

\section{Problem Statement} \label{sec:problem-statement}

We consider the problem of learning the high-fidelity dynamics of a nonlinear robotic system represented as a vector field of coupled, discrete-time nonlinear dynamics of the form:
\begin{equation}\label{eqn:dt-model}
  \begin{aligned}
    x_{k+1}^{\text{high}} &= f^{\text{high}}(x_k^{\text{high}}, u_k, \xi_k, \Delta x_{k}, \Delta k),
  \end{aligned}
\end{equation}
where $x_k^{\text{high}} \in \mathbb{R}^{n}$ are the high-fidelity, fused system states and $u_k \in \mathbb{R}^{m}$ are the control inputs. The vector $\xi_k \in \mathbb{R}^{o}$ denotes time-varying sensor noise present within onboard utilized sensors, while $\Delta x_k \in \mathbb{R}^{n}$ represents unmodeled disturbances acting on the system dynamics. The non-uniform time step $\Delta k \in \mathbb{R}$ captures discretization variability between successive states $x_k$ and $x_{k+1}$.

The estimation task is to approximate the true, high-fidelity dynamics $f^{\text{high}}$ under these uncertainties using only learned low-fidelity model $f^{\text{low}}_{\Gamma}$ and a learned residual function $\mathcal{R}_{\Gamma}(\cdot)$ between low-fidelity model outputs $\hat{x}_{k+1}^{\text{low}}$ and high-fidelity ground truth states $x_{k+1}^{\text{high}}$. To this end, we aim to learn a joint, high-fidelity surrogate model $f_{\Gamma}^{\text{high}}(x_k^{\text{low}}, u_k, \Delta k): \mathbb{R}^{n} \times \mathbb{R}^{m} \times \mathbb{R} \rightarrow \mathbb{R}^{n}$, parameterized by $\Gamma=\{\Gamma_{\text{low}},\Gamma_{\mathcal{R}}\}$ that provides reliable predictions of $x_{k+1}^{\text{high}}$ given noisy, partially observed inputs. 

To address the model mismatch problem inherent in purely data-driven approaches, we incorporate physics-based models $g_{i}(x_k,u_k,\Delta k): \mathbb{R}^{n} \times \mathbb{R}^{m} \times \mathbb{R} \rightarrow \mathbb{R}^{n}$ for $i=1,2$ fidelity levels, which captures system dynamics at different fidelity levels but still neglects the influence of sensor noise $\xi_k$ and unmodeled disturbances $\Delta x_k$. We task the low-fidelity model $f^{\text{low}}_{\Gamma}$ to learn the simplified dynamics found in $g_{1}(\cdot)$. Along with this, higher-fidelity model $g_{2}(\cdot)$ is used as a structural prior for the residuals $r_{k+1}$ predicted by $\mathcal{R}_{\Gamma}(x_{k}^{g_2},u_{k},\hat{r}_{k+1}):\mathbb{R}^{n}\times\mathbb{R}^{m}\times\mathbb{R}^{n}\rightarrow\mathbb{R}^{n}$. Following the residual multi-fidelity framework, the learned dynamics can be represented as:
\begin{equation}\label{eqn:residual-form}
    \begin{aligned}
        f_{\Gamma}^{\text{high}}&= f^{\text{low}}_{\Gamma}(x_k^{\text{low}}, u_k, \Delta k, \hat{x}_{k+1}^{\text{low}}) + \mathcal{R}_{\Gamma}(x_k^{g_2}, u_k, \hat{r}_{k+1}), \\
        r_{k+1} &= \hat{x}_{k+1}^{\text{low}}-x_{k+1}^{\text{high}}, \\
        \hat{r}_{k+1} &= \hat{x}_{k+1}^{\text{low}} - x_{k+1}^{g_{2}},
    \end{aligned}
\end{equation}
where $x_{k}^{\text{low}}\sim g_{1}(\cdot)$, $x_{k+1}^{g_2}\sim g_2(\cdot)$, and $r_{k+1}$ are true residual labels, and $\hat{r}_{k+1}$ are approximate physics-informed residual labels such that we assume $\hat{r}_{k+1}\approx\ r_{k+1}$. Regarding fidelity order, it should be understood that the following inequality holds: $\text{low}=g_{1}<g_2<\text{high}$. Unlike standard neural networks, neural processes provide predictive distributions conditioned on context-target sets, enabling the model to encode uncertainty about $\mathcal{R}_{\Gamma}(\cdot)$ and adapt to nonstationary dynamics. The problem of multi-fidelity residual state estimation thus reduces to learning $f_{\Gamma}^{\text{low}}$ and $\mathcal{R}_{\Gamma}(\cdot)$ such that $f^{\text{high}}\approx f_{\Gamma}^{\text{high}}$ while simultaneously ensuring calibrated predictive uncertainty over $f_{\Gamma}^{\text{high}}$ via an adaptive conformal prediction framework.

This formulation explicitly exploits the inductive bias of physics-based priors through $g_{1}(\cdot),\;g_{2}(\cdot)$ while retaining the flexibility of data-driven learning through the neural process backbone of this approach. In doing so, it enables the estimator to remain accurate in the presence of unmodeled sensor noise, nonlinear disturbances, and time-varying uncertainties---challenges that conventional dead-reckoning and single-fidelity learning approaches cannot robustly handle.

\section{Methodology}\label{sec:method}

\begin{figure*}[t]
    \centering
    \includegraphics[width=\textwidth]{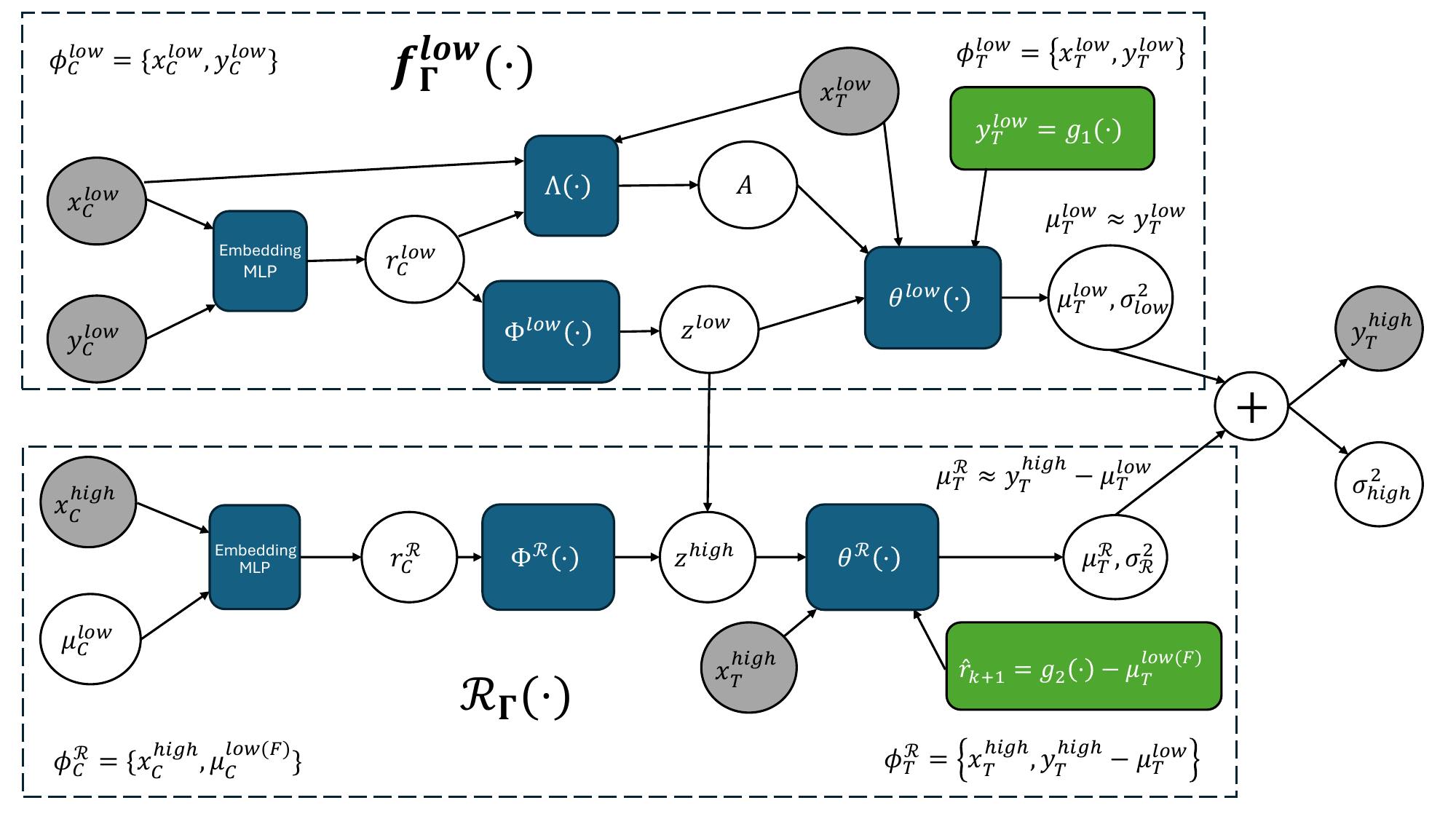} 
    \caption{Architecture and Direct Probabilistic Graph of MFR-PINP model where arrows show both model flow and conditional dependence. We display the notation for context $\phi_{C}^{\text{low}},\phi_{C}^{\mathcal{R}}$ and target $\phi_{T}^{\text{low}},\phi_{T}^{\mathcal{R}}$ sets utilized by both NP models. Note that shaded circles denote observed parameters, white denote learned parameters, blue denote utilized neural networks, and green represents physics-informed priors that are given to model decoders as input. We explicitly state that our model formulation builds on model structure discussed in \cite{niu2024multi} by incorporating physics-informed priors in model decoders for improved performance in modeling key conditional distributions $p(y_{T}^{\text{low}}|x_{T}^{\text{low}},z^{\text{low}},\hat{y}_{T}^{\text{low}},A)\;\;\text{and}\;\;p(r_{k+1}|x_{T}^{\text{high}},z^{\text{high}},\hat{r}_{k+1})$.}
    \label{fig:mfr-pinp-architecture}
\end{figure*}

\subsection{Model Architecture}
The flow chart in Figure \ref{fig:mfr-pinp-architecture} summarizes the full computational forward pass architecture of the Multi-Fidelity Residual Physics-Informed Neural Process (MFR-PINP) model utilized for our desired multi-fidelity, data-driven state estimation task. 

When analyzing the structure of both low-fidelity model $f_{\Gamma}^{\text{low}}(\cdot)$ and residual model $\mathcal{R}_{\Gamma}(\cdot)$, like other NP architectures, we utilize contextual sets $\phi_{C}^{\text{low}} =\{x_{C}^{\text{low}},y_{C}^{\text{low}}\}$ and $\phi_{C}^{\mathcal{R}}=\{x_{C}^{\text{high}},\mu_{C}^{\text{low}}\}$ along with queried inputs $x_{T}^{\text{low}},x_{T}^{\text{high}}$ to obtain predictions for queried outputs $y_{T}^{\text{low}}, y_{T}^{\text{high}}$. 

Within the estimation domain, we can consider $\left\{x_{C}^{\text{low}},x_{C}^{\text{high}}\right\}$ and $\left\{y_{C}^{\text{low}},\mu_{C}^{\text{low}}\right\}$ as contextual inputs and outputs for both $f_{\Gamma}^{\text{low}}(\cdot)$ and $\mathcal{R}_{\Gamma}(\cdot)$ model components. The contextual inputs may contain quantities such as an observed control action $u_k$ and other modeled external forces $\Delta \hat{x}_{k}$. Additionally, contextual outputs may consist of an estimate of the previous states generated by low-fidelity sensor input $x_{k}^{\text{low}}$ and mean predictions $\mu_{C}^{\text{low}}$ for $x_{k}^{\text{low}}$ generated from low-fidelity decoder $\theta^{\text{low}}(\cdot)$. It should be explicitly understood that within the residual physics-informed prior, $\mu_{C}^{\text{low}(F)}$ represents the low-fidelity mean prediction generated from a parameter-frozen low-fidelity decoder $\theta^{\text{low}(F)}(\cdot)$ that is periodically updated. This is done to ensure training stability when tasking the residual model to learn the gap between predictions from the consistently updated low-fidelity decoder and the ground truth. Therefore, a candidate definition for utilized context data in the MFR-PINP can be represented as:
\begin{align*}
    x_{C}^{\text{low}} &= \{u_{k}\},\;\; 
    x_{C}^{\text{high}} = \{u_{k}, \Delta \hat{x}_{k}\} \\
    y_{C}^{\text{low}} &= \{x_{k}^{\text{low}}\},\;\; 
    y_{C}^{\mathcal{R}} = \{\mu_{C}^{\text{low}}\}
\end{align*}  
where $y_{C}^{\mathcal{R}}$ is simply a placeholder to define the residual contextual output. Similarly, we can define target inputs $x_{T}^{\text{low}},x_{T}^{\text{high}}$ and outputs $y_{T}^{\text{low}},y_{T}^{\text{high}}$ as data that encodes information about the desired next state estimate. As such, we can define low and high fidelity target sets as the following:

\begin{align*}
    x_{T}^{\text{low}} &= \{\Delta k\},\;\; 
    x_{T}^{\text{high}} = \{\Delta k\} \\
    y_{T}^{\text{low}} &= \{x_{k+1}^{\text{low}}\},\;\;
    y_{T}^{\text{high}} = \{x_{k+1}^{\text{high}}\}
\end{align*}

where $\Delta k$ represents the (possibly non-uniform) time step between previous states and next states. In this way, we assume $x_{T}^{\text{low}}=x_{T}^{\text{high}}$; however, depending on the desired system, further information that provides \emph{future-state} specific input data could be included in $x_{T}^{\text{high}}$.

In observing the intricate steps within the MFR-PINP's forward computation, we note that context set $\phi_{C}^{\text{low}}$ is sent through an embedding multi-layer perception (MLP) to obtain permutation-invariant bottleneck representations $r_{C}^{\text{low}},r_{C}^{\text{high}}$. Those computed representations $r_{C}^{\text{low}},r_{C}^{\text{high}}$ are used in the generation of latent samples $\{z^{\text{low}},z^{\text{high}}\}$ through variational encoders $\Phi^{\text{low}}(\cdot),\Phi^{\mathcal{R}}(\cdot)$ which parameterize latent Gaussian distributions. These latent samples capture hidden features from our context sets and encode global uncertainty in predictions formed in the decoder-generated state predictions. It should also be known that we utilize a hierarchical latent structure in determining high-fidelity latent sample $z^{\text{high}}$ where information from lower fidelity sample $z^{\text{low}}$ is included to further encourage for more robust cross-fidelity learning for the residual model $\mathcal{R}_{\Gamma}(\cdot)$.

Note that to improve expressibilty in the low-fidelity architecture $f_{\Gamma}^{\text{low}}$, we incorporate a deterministic path network $\Lambda(\cdot)$ which takes as input keys ($x_{C}^{\text{low}}$), values ($r_{C}^{\text{low}}$), and queries ($x_{T}^{\text{low}}$) then computes a cross-attention matrix $A$ that is fed to the low-fidelity decoder as an additional conditioning parameter which prevents a context-underfitting phenomena caused by the computations of bottlenecks $r_{C}^{\text{low}},r_{C}^{\text{high}}$. Additionally, we implement a self-attention mechanism for the low-fidelity model between low-fidelity contextual input $x_{C}^{\text{low}}$ and output $y_{C}^{\text{low}}$ to encourage further understanding between the state dynamics found in $y_{C}^{\text{low}}$ and the corresponding effects of system inputs found in $x_{C}$.

Each model's decoder $\Theta^{\text{low}}(\cdot),\Theta^{\mathcal{R}}(\cdot)$ receives their target inputs $x_{T}^{\text{low}},x_{T}^{\text{high}}$ and other represented conditional parameters to generate predicted target mean $\mu_{T}^{\text{low}},\mu_{T}^{\mathcal{R}}$ and variance $\sigma_{low}^2,\sigma_{\mathcal{R}}^2$. Observe that these quantities predicted by the decoder are, then, summed together to form the combined model predictions of the form: 

\begin{equation}
\begin{aligned}
    \mu^{\text{high}}_{T} &=
    \mu_{T}^{\text{low}}+\mu_{T}^{\mathcal{R}},\;\;\;
    \sigma^{2}_{\text{high}} =  \sigma_{\text{low}}^{2}+\sigma_{\mathcal{R}}^{2},
\end{aligned}
\end{equation}
where we desire $\mu_{T}^{\text{high}}\approx y_{T}^{\text{high}}$. As mentioned prior, our MFR-PINP formulation builds on the high-fidelity surrogate modeling done by the MFRNP through integrating physics-informed priors at multiple fidelity levels to achieve lower prediction error and obtain predictions that are more \emph{physically consistent} than purely data-driven methods. At the lowest fidelity, a simplified physics model $g_{1}(x_{k}, u_{k}, \Delta k)$ represents baseline kinematics, such as wheel odometry, that neglect noise $\xi_{k}$ and disturbances $\Delta x_{k}$. A higher-fidelity structural prior $g_{2}(x_{k}, u_{k}, \Delta k)$ such as one with identified plant parameters encodes more accurate but still approximate dynamics through extended kinematic modeling or higher-order approximations.

The low-fidelity neural process model $f^{\text{low}}_{\Gamma}(x_{k}^{\text{low}}, u_{k}, \Delta k)$ learns to reproduce the dynamics of $g_{1}(\cdot)$ given low-fidelity motion data. To bridge the gap between low-fidelity predictions and the true high-fidelity system, we utilize a residual neural process $\mathcal{R}_{\Gamma}(\cdot)$ that models the correction relative to $x_{k+1}^{\text{high}}$. The final, fused estimates of the predicted target dynamics $\mu_{T}^{\text{high}}\in\mathbb{R}^{n}$ and their state standard deviations $\sigma_{\text{high}}\in\mathbb{R}^{n}$ are represented as:
\begin{equation}
\begin{aligned}
    \mu^{\text{high}}_{T} &=
    f^{\text{low}}_{\Gamma}(x^{\text{low}}_{k}, u_{k}, \Delta k, x_{k+1}^{\text{low}}) + \mathcal{R}_{\Gamma}(x^{g_2}_{k}, u_{k},\Delta k, \hat{r}_{k+1}), \\
    \sigma_{\text{high}} &=  \sqrt{\sigma_{\text{low}}^{2}+\sigma_{\mathcal{R}}^{2}}
\end{aligned}
\end{equation}
with true and approximate residual labels defined, respectively, as:
\begin{equation}
    \begin{aligned}
         r_{k+1} &= \mu_{T}^{\text{low}(F)} - x^{\text{high}}_{k+1},\;\; 
         \hat{r}_{k+1} = \mu_{T}^{\text{low}(F)} - x^{g_2}_{k+1},
    \end{aligned}
\end{equation}
where $x^{\text{low}}_{k},x_{k+1}^{\text{low}} \sim g_{1}(\cdot)\;\text{ and } x^{g_{2}}_{k+1} \sim g_{2}(\cdot)$. By explicitly leveraging fidelity ordering low $=g_{1}< g_{2} <$ high, the MFR-PINP fuses multi-fidelity priors with learned residuals to approximate the true dynamics $f^{\text{high}}$.

\subsection{Model Optimization}

The training objective combines separate ELBO (evidence lower bound) losses for low-fidelity model $f_{\Gamma}^{\text{low}}$ and residual model $\mathcal{R}_{\Gamma}$ that are optimized jointly. These dual ELBO terms are included in the following joint variational lower-bound (ELBO) optimization:
    \[\Gamma_{\text{low}}^{*}, \Gamma_{\mathcal{R}}^{*}  = \arg\max_{\Gamma^{*}} \mathcal{L}^{\text{low}} + \mathcal{L}^{\mathcal{R}},
    \] \label{eqn:model-optimization}
    subject to
\begin{equation}
\begin{aligned}
\log p(y_{T}^{\text{low}}|x_{T}^{\text{low}}, \phi_{C}^{\text{low}}) &\geq \mathcal{L}^{\text{low}}(p_{\Theta}^{\text{low}},q_{\Phi}^{\text{low}}, A) \\
&= \log p_{\Theta}^{\text{low}}(y_{T}^{\text{low}}|x_{T}^{\text{low}},z^{\text{low}}, \hat{y}_{T}^{\;\text{low}},A) \\
&- \log \frac{q_{\Phi}^{\text{low}}(z^{\text{low}}|\phi_{C}^{\text{low}})}{q_{\Phi}^{\text{low}}(z^{\text{low}}|\phi_{T}^{\text{low}})}
\end{aligned}
\end{equation}

\begin{equation}
\begin{aligned}
    \log p(r_{k+1}|x_{T}^{\text{high}}, \phi_{C}^{\text{high}}) &\geq \mathcal{L}^{\mathcal{R}}(p_{\Theta}^{\mathcal{R}},q_{\Phi}^{\mathcal{R}}) \\ 
    &= \log p_{\Theta}^{\mathcal{R}}(r_{k+1}|x_{T}^{\text{high}},z^{\text{high}},\hat{r}_{k+1}) \\
    &- \log \frac{q_{\Phi}^{\mathcal{R}}(z^{\text{high}}|z^{\text{low}}, \phi_{C}^{\mathcal{R}})}{q_{\Phi}^{\mathcal{R}}(z^{\text{high}}|z^{\text{low}}, \phi_{T}^{\mathcal{R}})}
\end{aligned}
\end{equation}
where $\mathcal{L}^{\text{low}}, \mathcal{L}^{\mathcal{R}}$ are low-fidelity and residual loss terms, respectively. Optimization is performed with an Adam solver at each step using mini batches of context-target trajectories. The NP latent variables allow amortized inference, enabling fast adaptation of the residual model across diverse conditions without retraining from scratch. To obtain more information regarding the MFR-PINP's online learning and inference procedures, observe the contents found in Algorithm \ref{alg:mfrpinp-training-inference-alg}.

When observing Algorithm \ref{alg:mfrpinp-training-inference-alg}, it is important to note that $\mathcal{D}_{rb}^{\text{low}},\mathcal{D}_{rb}^{\text{high}}$ represent the replay buffer datasets that contain low and high fidelity samples, respectively. We start the MFR-PINP training with $|\mathcal{D}_{rb}^{\text{low}},\mathcal{D}_{rb}^{\text{high}}|=10^5$ samples each with per-training iteration batch size $|\mathcal{D}_{j}^{\text{low}},\mathcal{D}_{j}^{\text{high}}|=2000$ to mitigate the slow initial convergence issues of online training \cite{kaushik2020online}. Additionally, we denote $q_{\alpha}$ as the quantile used to scale the MFR-PINP's predicted high-fidelity standard deviation $\sigma_{\text{high}}$ to ensure that the true high-fidelity states fall within the model's uncertainty bounds at a user-defined rate $\alpha\in[0,1]$. For more information regarding the split conformal method utilized to gain quantified uncertainty for our estimation method, refer to the following subsection \ref{subsec:conformal-method}. Note that for decoder models $p_{\Theta}^{\text{low}},\;p_{\Theta}^{\mathcal{R}},\;\text{and} \;\Theta^{\text{low}(F)}$, we utilize shorthand notation for their conditional arguments, but they can be observed here \eqref{eqn:model-optimization}. For optimization, we utilize the Adam optimizer with learning rate $lr=10^{-3}$ and weight decay regularization parameter $\lambda=5\times10^{-7}$ to ensure fast and stable training convergence.

\begin{algorithm}
\caption{MFR-PINP Hybrid Training \& Inference Algorithm}
\label{alg:mfrpinp-training-inference-alg}
\begin{algorithmic}[1]

\Require $\Gamma_{0} = \{\Gamma_{\text{low}(0)}, \Gamma_{\mathcal{R}(0)}\},\;
         \mathcal{D}_{rb}^{\text{low}}, \mathcal{D}_{rb}^{\text{high}}, \mathcal{D}_{0}^{\text{low}},\mathcal{D}_{0}^{\text{high}}
         \vspace{0.2em}$
\Statex \hspace{\algorithmicindent}$\quad\; g_{1}(\cdot),\; g_{2} (\cdot),\;q_{\alpha}$
\While{$isRunning = \text{True}$}
    \State \textbf{MFR-PINP Training Phase} (Executed periodically)
    \State Sample random batch $\mathcal{D}_{j}^{\text{low}}, \mathcal{D}_{j}^{\text{high}}\sim\mathcal{D}_{\text{rb}}^{\text{low}}, \mathcal{D}_{\text{rb}}^{\text{high}}$
    \State \text{Parse Batch LF Dataset $\mathcal{D}_{j}^{\text{low}}$ into $\phi_{C}^{\text{low}},\;\phi_{T}^{\text{low}}$}
    \State \text{Sample $z^{\text{low}} \sim q_{\Phi}^{\text{low}}(z^{\text{low}}|\phi_{T}^{\text{low}})$}
    \State \text{Compute Attention Matrix $A\gets\Lambda_{i}(x_{C}^{\text{low}},r_{C}^{\text{low}},x_{T}^{\text{low}})$}
    \State \text{Compute LF state prior $y_{T}^{\text{low}}\gets g_1(\cdot)$}
    \State \text{Predict LF states $\mu_{T}^{\text{low}},\sigma_{\text{low}}^{2} \gets p_{\Theta}^{\text{low}}(y_{T}^{\text{low}}|\cdot)$}
    \State \text{Compute target $\mu_{T}^{\text{low}(F)}\gets \Theta^{\text{low}(F)}(y_{T}^{\text{low}}|\cdot)$}
    \State \text{Sample $z^{\text{high}}\sim q_{\Phi}^{\mathcal{R}}(z^{\text{high}}|z^{\text{low}},\phi_{T}^{\text{high}})$}
    \State \text{Parse Batch HF Dataset $\mathcal{D}_{j}^{\text{high}}$ into $\phi_{C}^{\mathcal{R}},\phi_{T}^{\mathcal{R}}$}
    \State \text{Compute true residual $r_{k+1} \gets y_{T}^{\text{high}} - \mu_{T}^{\text{low}(F)}$}
    \State \text{Compute approximate residual $\hat{r}_{k+1}\gets x_{k+1}^{g_2}-\mu_{T}^{\text{low}(F)}$}
    \State \text{Predict residuals $\mu_{T}^{\mathcal{R}},\sigma_{\mathcal{R}}^2 \gets p_{\Theta}^{\mathcal{R}}(r_{k+1}|\cdot)$}
    \State \text{Backpropogate by using optimization found in \eqref{eqn:model-optimization}}
    \vspace{0.2cm}
    \State \textbf{MFR-PINP Inference Phase} (Executed per iteration)
     \State \text{Parse LF sample $\mathcal{D}_{i}^{\text{low}}$ into $\phi_{C}^{\text{low}},\;\phi_{T}^{\text{low}}$}
    \State \text{Sample $z^{\text{low}} \sim q_{\Phi}^{\text{low}}(z^{\text{low}}|\phi_{C}^{\text{low}})$}
    \State \text{Compute Attention Matrix $A\gets\Lambda_{i}(x_{C}^{\text{low}},r_{C}^{\text{low}},x_{T}^{\text{low}})$}
    \State \text{Compute LF state prior $y_{T}^{\text{low}}\gets g_1(\cdot)$}
    \State \text{Predict LF states $\mu_{T}^{\text{low}},\sigma_{\text{low}}^{2}\gets p_{\Theta}^{\text{low}}(y_{T}^{\text{low}}|\cdot)$}
    \State \text{Sample $z^{\text{high}}\sim q_{\Phi}^{\mathcal{R}}(z^{\text{high}}|z^{\text{low}},\phi_{C}^{\text{high}})$}
    \State \text{Compute approximate residual $\hat{r}_{k+1}\gets x_{k+1}^{g_2}-\mu_{T}^{\text{low}}$}
    \State \text{Predict residuals $\mu_{T}^{\mathcal{R}},\sigma_{\mathcal{R}}^2 \gets p_{\Theta}^{\mathcal{R}}(r_{k+1}|\cdot)$}
    \State \text{Compute HF fused estimates $\mu_{T}^{\text{high}}=\mu_{T}^{\text{low}}+\mu_{T}^{\mathcal{R}}$ and}
    \Statex $\;\;\;\;\;\sigma_{\text{high}}^{2}=\sigma_{\text{low}}^{2}+\sigma_{\mathcal{R}}^{2}$
    \State $\text{Apply CP quantile to std deviation}$  
    \Statex $\quad\quad\sigma_{\text{high}}\gets \sqrt{\sigma_{\text{high}}^{2}}\cdot q_{\alpha}$
    \vspace{0.2cm} 

    \State \textbf{Data Updates/Frozen Model Updates} \\ $\;\;\;\;\;$(Executed based on user-defined rates)
    \State \text{New LF observed data: $\mathcal{D}_{i}^{\text{low}}\gets \mathcal{D}_{i+1}^{\text{low}}$  (per iter)}
    \State \text{New HF observed data: $\mathcal{D}^{\text{high}}_{i}\gets \mathcal{D}_{i+1}^{\text{high}}$   (per few iters)}
    \State \text{Update LF Target: $\Theta_{i+1}^{\text{low}(F)}\gets \Theta_{i+1}^{\text{low}}$}  (per $\sim100$ iters)
    \State $\text{Append}\;\mathcal{D}_{rb}^{\text{low}}\gets \mathcal{D}_{i}^{\text{low}}\;\text{and}\;\mathcal{D}_{rb}^{\text{high}}\gets \mathcal{D}_{i}^{\text{high}}$
    \State $\text{Remove oldest samples: }\mathcal{D}_{rb}^{\text{low}}.pop(-1), \mathcal{D}_{rb}^{\text{high}}.pop(-1) $
    \State $i \gets i + 1$
\EndWhile

\end{algorithmic}
\end{algorithm}

For deployment, the MFR-PINP operates in real time using primarily fast, low-fidelity sensor inputs (i.e. IMU, encoders) with infrequent high-fidelity residual updates provided by more expensive sensor inputs (i.e. camera, LiDAR). During inference, the model conditions on the most recent context window to generate predictive distributions for $x_{k+1}^{\text{high}}$. The residual process $R_{\Gamma}$ dynamically adjusts corrections based on discrepancies between $y_{T}^{\text{high}}$ and the low-fidelity prediction $\mu_{T}^{\text{low}}$ iteratively.

\subsection{Split Conformal Prediction for Uncertainty Guarantees} \label{subsec:conformal-method}
To ensure reliable uncertainty estimates under real-world variability, we integrate a split conformal scheme that updates noncomformity quantiles $q_{\alpha}$ based on batches of most recent sensor data \cite{barber2023conformal}. Periodically, conformal calibration is applied to the most recent prediction residuals to generate uncertainty sets with the following finite-sample coverage guarantees for next incoming sensor data $(x_{n+1}, y_{n+1})$:
\begin{equation}
\begin{aligned}
    p(y_{n+1}\in\mathcal{C}(x_{n+1})) &\geq 1 - \alpha \\ 
    p \left(|y_{n+1}-\hat{y}_{n+1}| \leq \sqrt{q_{\alpha}\cdot\hat{\sigma}_{n+1}}\right) &\geq 1-\alpha,
\end{aligned}
\end{equation}
where $\alpha\in[0,1]$ is user-defined and $\hat{y}_{n+1},\hat{\sigma}_{n+1}$ are the predicted mean and standard deviation of the model, respectively. This yields intervals that adapt to time-varying sensor noise, unmodeled disturbances, and environmental changes.

\section{Results \& Validation} \label{sec:results}

In this section, we validate the MFR-PINP estimation approach by evaluating its performance on a physical robotic platform for real-time usage. We also verify that our approach augmented with conformal quantized regression enhances uncertainty quantification by improving the distribution modeling capabilities of the MFR-PINP over the baseline models during both training and inference through several evaluation metrics. Recall that the baselines by which we compare the MFR-PINP's performance are the unscented Kalman filter (UKF) and transformer-based deep Kalman filter. Our deep Kalman filter (DKF) implementation utilizes two decoder-only transformers that model the process and measurement models used in the linear Kalman filter formulation. Due to the use of neural network modeling, however, nonlinear behavior from either model can be captured within the DKF. Pertaining to the UKF, note that this is the utilized ground-truth (oracle) for this study due to high-fidelity nature of sensor fusion approaches. For more details regarding the UKF implementation, refer here \ref{subsec:robot-ground-truth-design}.

\subsection{Experimental Robotic Platform Used in Study}

\begin{figure}[h]
    \centering
    \includegraphics[height=5cm,width=5cm]{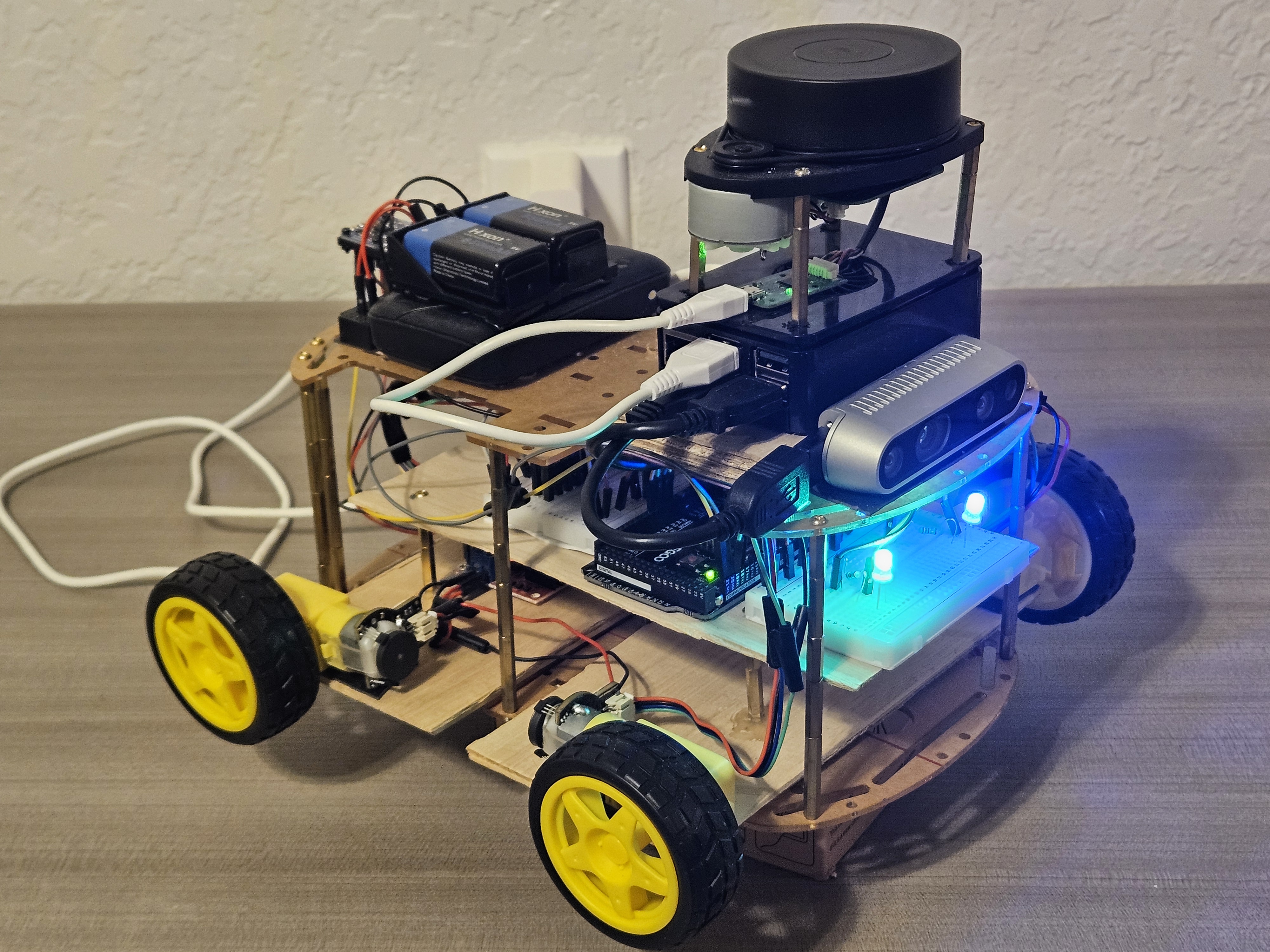}
    \caption{Four-wheeled robotic platform used in study}
    \label{fig:robot-pic}
\end{figure}

As can be observed in Figure \ref{fig:robot-pic}, the robotic platform provided for our study is a four-wheeled skid-steering robot that is equipped with a Raspberry Pi 5 (8 GB RAM) as the central unit, wheel encoders, a 9-DOF IMU that is equipped with a gyroscope, accelerometer, and magnetometer, a 360-degree 2D LiDAR, and D435i realsense camera. We define low-fidelity sensors as the encoders and IMU since these are pure motion sensors that do not require knowledge of external landmarks to localize; however, that lack of landmark correction leads to sensor drift and suboptimal odometry estimates\cite{niu2021wheel}. Conversely, we define the high fidelity odometry sources as those being generated from the LiDAR and depth camera due to the improved drift correction measures that they provide. However, due to the heavy-computation nature of these odometry sources, they are included more sparingly in odometry updates than low-fidelity sources.

\subsection{Dynamics Models Used for Skid-Steering Robot} \label{subsec:robot-dynamic-models}

As mentioned in both the Problem Statement \ref{sec:problem-statement} and the methodology \ref{sec:method}, the MFR-PINP requires access to two dynamic models $g_1(\cdot)$ and $g_2(\cdot)$ for fidelity levels $k=1,2$. For low-fidelity model $g_1(\cdot)$, we assume a standard differential drive kinematic model for a planar 2D environment since the dynamics of this four-wheeled robot can be assumed similar. Therefore, for planar positional states $x^I,y^I,\theta^I$ in some inertial frame $I$, we represent the dynamics of $g_1(\cdot)$ as the following:

\begin{equation}
    \begin{bmatrix}
        \dot{x}^I \\
        \dot{y}^I \\
        \dot{\theta}^I
    \end{bmatrix} = \begin{bmatrix}
        (r_{w}/2)\cos(\theta^I) & (r_{w}/2)\cos(\theta^I)  \\
        (r_{w}/2)\sin(\theta^I) & (r_{w}/2)\sin(\theta^I) \\
        -r_{w}/L & r_{w}/L 
    \end{bmatrix} \begin{bmatrix}
        \omega_R \\
        \omega_L
    \end{bmatrix},
\end{equation}
where $r_{w}$ is the robot wheel radius, $L$ is the horizontal length of the robot chassis from wheel-to-wheel, and $\omega_{R},\omega_{L}$ are the left and right wheel velocities, respectively. We assume that the velocity difference between either the left and right set of wheels are negligible to use the above dynamics model.

To model the higher fidelity dynamics $g_{2}(\cdot)$, we utilize a planar Newton-Euler skid-steering model that incorporates linear slippage dynamics and inertial coupling. This model incorporates second-order dynamics through the inclusion of longitudinal $u^b$ and lateral $v^b$ velocities in the robot's body frame, then relates these velocities along with yaw rate $\dot{\theta}^I$ in the following higher-fidelity dynamics model:
\begin{equation}
    \begin{aligned}
        F_{x}&=\frac{C_{t}}{2}(\omega_{R}+\omega_{L}),\; F_{y}=-C_{\alpha}v^{b},\; M_{z}=LC_{t}(\omega_{R}-\omega_{L})\\
        \dot{u}^b &= \frac{F_{x} -mv^{b}\dot{\theta}^I}{m},\;\; \dot{v}^b = \frac{F_{y}-mu^{b}\dot{\theta}^I}{m},\;\; 
        \ddot{\theta}^I = \frac{M_{z}}{I_{z}}, \\[0.2cm]
        \begin{bmatrix}
            \dot{x}^I \\
            \dot{y}^I
        \end{bmatrix} &= \begin{bmatrix}
            \cos(\theta^I) & -\sin(\theta^I) \\
            \sin(\theta^I) & \cos(\theta^I)
        \end{bmatrix}\begin{bmatrix}
            u^b \\
            v^b
        \end{bmatrix},
    \end{aligned}
\end{equation}

where $F_{x}$ is the net longitudinal force, $F_{y}$ is the lateral slip force, and $M_{z}$ is the yaw moment about the robot's center of mass. Additionally, $C_{t}$ is the longitudinal traction gain and $C_{\alpha}$ is the corresponding lateral slip gain. All static kinematic parameters utilized by both $g_{1}(\cdot)$ and $g_{2}(\cdot)$ for our designed robot were determined through extensive experimental analysis beforehand. A table that summarizes the determined values of each parameter for our robot is located in Table \ref{tab:kinematic-properties}.  

\begin{table}[h!]
  \centering
  \scriptsize  
  \begin{tabular}{|c|c|c|c|c|c|}
    \hline
    $m$ (kg) & $r_w$ (m) & $L$ (m) & $I_z$ (kg$\cdot \text{m}^2$) & $C_t$ (N$\cdot$s/m) & $C_\alpha$ (N$\cdot$s/m)\\
    \hline
    10.7 & 0.034 & 0.288 & 4.35 & 15.0 & 11.5 \\
    \hline
    
  \end{tabular}
  \caption{Kinematic Properties of Robotic Platform}
  \label{tab:kinematic-properties}
\end{table}

\subsection{Context/Target Formulation for MFR-PINP Usage}
As stated within the Methodology \ref{sec:method}, the MFR-PINP utilizes both low-fidelity and residual context and target $\phi_{T}^{\text{low}},\phi_{T}^{\mathcal{R}}$ sets in its prediction scheme. Specifically, both low-fidelity model and residual model access context sets $\phi_{C}^{\text{low}},\phi_{C}^{\mathcal{R}}$ along with target inputs $x_{T}^{\text{low}},x_{T}^{\text{high}}$ to infer the values of target outputs $y_{T}^{\text{low}},y_{T}^{\text{high}}$. Pertaining to usage for the designed robot with system states $x^{I},y^{I},\theta^{I}$ and their time derivatives $\dot{x}^{I},\dot{y}^{I},\dot{\theta}^{I}$, we exploit the following relationship between low-fidelity/high-fidelity sensor inputs:

\begin{equation}
    \begin{aligned}
        x_{C}^{\text{low}}&=\{\omega_{R},\omega_{L}\},\;x_{C}^{\text{high}} = \{\omega_{R},\omega_{L}, F_{x},F_{y},M_{z}\} \\
        y_{C}^{\text{low}}&=\{x_{k}^{\text{low}}\},\; y_{C}^{\mathcal{R}} = \{\mu_{C}^{\text{low}}\} \\
        x_{T}^{\text{low}}&=\{\Delta k\},\; x_{T}^{\text{high}} = \{\Delta k\}\\
        y_{T}^{\text{low}}&=\{x_{k+1}^{\text{low}}\}\;, y_{T}^{\text{high}} = \{x_{k+1}^{\text{high}}\},
    \end{aligned}
\end{equation}

where $x_{k}^{\text{low}},x_{k+1}^{\text{low}}\sim g_{1}(\cdot)$ from \ref{subsec:robot-dynamic-models}, $\Delta k$ represents the time interval between states $x_{k}$ and $x_{k+1}$. We denote the state vector of this system as $x_{k}=\begin{bmatrix}
    x^{I}_{k} & y^{I}_{k} & \theta^{I}_{k} & \dot{x}^{I}_{k} & \dot{y}^{I}_{k} & \dot{\theta}^{I}_{k}
\end{bmatrix}$. Additionally, $\mu_{C}^{\text{low}}$ represents the generated low-fidelity mean prediction for previous low-fidelity state $x_{k}^{\text{low}}$ from low-fidelity model $f_{\Gamma}^{\text{low}}$ and $x_{k+1}^{\text{high}}$ is generated from the ground-truth discussed in \ref{subsec:robot-ground-truth-design}.

\subsection{Design of Ground Truth} \label{subsec:robot-ground-truth-design}

Pertaining to the high-fidelity ground-truth $y_{T}^{\text{high}}$ utilized in this study, we utilize the fused states from an unscented Kalman filter (UKF) that outputs odometry estimates at 50 Hz. This UKF processes sensor information from both the low-fidelity encoder and IMU signals (both publishing signals at 50 Hz) in addition to landmark odometry information provided from an onboard LiDAR and camera. Specifically, pertaining to the odometry sources generated from the LiDAR and camera, we utilize rtabmap (Real-Time Appearance-Based Mapping) nodes that compute stereo odometry at 5 Hz and ICP odometry at 10 Hz. We note that the UKF algorithm utilized to generate the ground-truth, high-fidelity data comes from the UKF node in the ROS2 \textit{robot\_localization} package \cite{moore2015generalized}.

\subsection{Error Metrics Used in Study}
To objectively compute the prediction error consistently with the related literature \cite{revach2022kalmannet,hurwitz2024deep}, we use the root mean squared error (RMSE) between the six states of the tracked robot $x^{I},y^{I},\theta^{I},\dot{x}^{I},\dot{y}^{I},\dot{\theta}^{I}$. The formula for RMSE is given by:
\begin{align*}
    \text{RMSE} = \sqrt{\frac{1}{N} \sum_{i=1}^{N} \sum_{j=1}^{6} \left(Y_{i,j} - \hat{Y}_{i,j}\right)^{2}},
\end{align*}
where $N$ is the batch size, and $Y, \hat{Y}\in \mathbb{R}$ are the true and predicted system states, respectively.
In addition to the above metric, we utilize negative log-likelihood (NLL) as a metric to compare the capabilities of modeling marginal distribution $p(y_{T}^{\text{high}}|x_{T}^{\text{high}},\phi_{C}^{\text{high}})$ between the CP-augmented MFR-PINP and other baselines. This metric is consistent with related literature in distribution modeling evaluation to ensure effects of distribution under/overconfidence are reflected \cite{kim2019attentive,revach2022kalmannet}. The formula for NLL is given by:
\begin{align*}
    \text{NLL} = \frac{1}{2N}\sum_{i=1}^{N}\sum_{j=1}^{6} \left(\log(2\pi \hat{\sigma}_{i,j}^{2}) + \frac{(Y_{i,j}-\hat{Y}_{i,j})^{2}}{\hat{\sigma}_{i,j}^{2}} \right)
\end{align*}

\subsection{Convergence Dynamics \& Test Trajectory Evaluations}

\begin{figure}[h]
    \centering
    \includegraphics[scale=0.50]{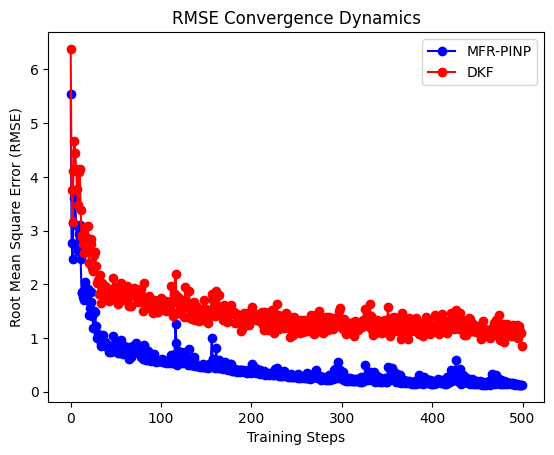}
    \caption{Model training dynamics of the MFR-PINP and transformer-based deep Kalman filter (DKF) \cite{goel2024can}}
    \label{fig:converge-compare}
\end{figure}

\begin{figure}[h]
    \centering
    \includegraphics[scale=0.50]{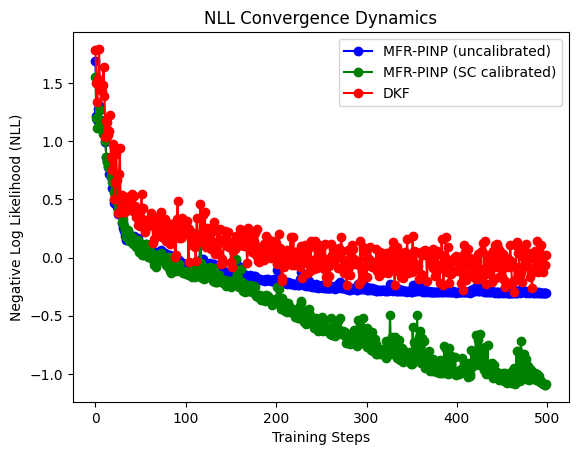}
    \caption{Distribution error dynamics of the split conformal method-augmented MFR-PINP (SC calibrated), the non-split conformal MFR-PINP (uncalibrated), and the deep Kalman filter (DKF)}
    \label{fig:nll-compare}
\end{figure}

\begin{figure}[h]
    \centering
    \includegraphics[scale=0.40]{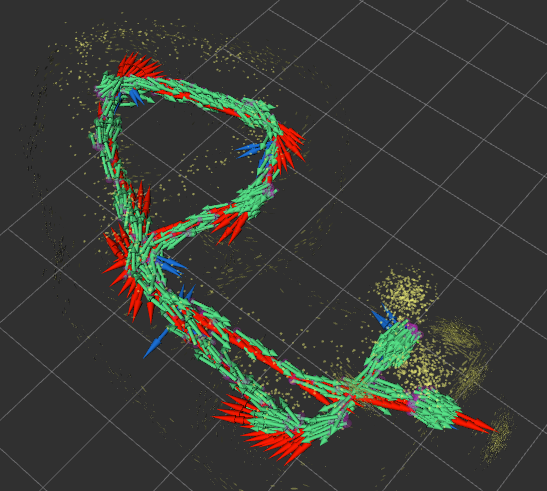}
    \caption{Visual results on ground-truth test trajectory (green) from MFR-PINP (blue) and DKF \cite{goel2024can} (red).}
    \label{fig:test-traj}
\end{figure}

When observing the convergence plots in Figures \ref{fig:converge-compare} and \ref{fig:nll-compare} along with Tables \ref{tab:train-RMSE-results} and \ref{tab:train-NLL-results}, one can see that the MFR-PINP outperforms the state-of-the-art transformer-based DKF \cite{goel2024can} in both raw prediction error (RMSE) and distributional modeling (NLL). Furthermore, when augmenting the MFR-PINP's uncertainty quantification with the split conformal method, it is determined from the NLL plot that the model gains deeper capabilities in modeling the target, high-fidelity dynamical distribution of the robot $p(y_{T}^{\text{high}}|x_{T}^{\text{high}}, \phi_{C}^{\text{high}})$ than both the uncalibrated version of the MFR-PINP in addition to the DKF. In our real-time testing application, we teleoperated the robot to follow a given test trajectory (seen in Figure \ref{fig:test-traj}) and recorded the formed trajectories from the MFR-PINP, DKF, and ground-truth UKF. As is seen in the accompanying Table \ref{tab:test-traj-results}, the MFR-PINP outperforms the DKF in tracking the UKF oracle trajectory by a significant margin. Each model was ran and output predictions at 50 Hz - showing viable real-time feasibility of all approaches.  

\begin{table}[h!]
  \centering
  \scriptsize  
  \begin{tabular}{|c|c|c|}
    \hline
     & MFR-PINP & DKF \\
    \hline
    Lowest RMSE & 0.154 & 0.958  \\
    \hline
    
  \end{tabular}
  \caption{Lowest RMSE reached for all models in Figure \ref{fig:converge-compare}}
  \label{tab:train-RMSE-results}
\end{table}

\begin{table}[h!]
  \centering
  \scriptsize  
  \begin{tabular}{|c|c|c|c|}
    \hline
     & SC calibrated & Uncalibrated & DKF \\
    \hline
    Lowest NLL & -1.274 & -0.201 & -0.192 \\
    \hline
    
  \end{tabular}
  \caption{Lowest NLL reached for all models in Figure \ref{fig:nll-compare}}
  \label{tab:train-NLL-results}
\end{table}

\begin{table}[h!]
  \centering
  \scriptsize  
  \begin{tabular}{|c|c|c|}
    \hline
     & MFR-PINP & DKF \\
    \hline
    RMSE & 1.164 & 2.015 \\
    \hline
    
  \end{tabular}
  \caption{RMSE of models on test trajectory in Figure \ref{fig:test-traj}}
  \label{tab:test-traj-results}
\end{table}

\section{Conclusions \& Future Works} \label{sec:conclude+future}
Through observation of the MFR-PINP's estimation results within the provided analyses compared to other model baselines, it has been shown that the MFR-PINP provides competitive performance in comparison to the state-of-the-art.  As a future research direction, we wish to improve the data efficiency of our hybrid training approach by introducing \emph{synthetic} sensor data samples during MFR-PINP training. By doing this, we mitigate the need for current, potentially-expensive data acquisition methods and address the well-known robotics sim2real gap for improved model performance. Additionally, we also wish to incorporate a learned measurement model $\hat{h}(\cdot)$ to provide more robust error correction for predicted states. Finally, as a proposed real-time approach, our developed CP-based uncertainty quantification method must be leveraged \emph{adaptively} throughout a mission, especially for safety-critical applications where we require the estimation error to be bounded through environmental changes. Beyond the single-agent robot state estimation task explored in this study, this novel, real-time, safety-critical MFR-PINP model formulation could be extended to a wider range of applications such as multi-agent estimation, model-based reinforcement learning, and model predictive control.  

\bibliographystyle{unsrt}
\bibliography{sources-real-time-NP}

\end{document}